\documentclass[conference]{IEEEtran}
\usepackage{cite}
\usepackage{amsmath,amssymb,amsfonts}
\usepackage{algorithmic}
\usepackage{graphicx}
\usepackage{textcomp}
\usepackage[dvipsnames]{xcolor}
\def\BibTeX{{\rm B\kern-.05em{\sc i\kern-.025em b}\kern-.08em
    T\kern-.1667em\lower.7ex\hbox{E}\kern-.125emX}}
\usepackage{makecell}

\newcommand{\rian}[1]{{\color{black}#1}}
\usepackage{tikz}
\usepackage{xcolor}
\newcommand*\circled[1]{\tikz[baseline=(char.base)]{
            \node[shape=circle,fill,inner sep=1pt] (char) {\textcolor{white}{#1}};}}
\usepackage{listings}
\usepackage{multirow}
\usepackage{url}

\begin{document}

\title{NAPER: Fault Protection for Real-Time Resource-Constrained Deep Neural Networks}

\author{\IEEEauthorblockN{Rian Adam Rajagede$^{\dagger 1}$, Muhammad Husni Santriaji$^{*2}$, Muhammad Arya Fikriansyah$^{3}$ \\ Hilal Hudan Nuha$^{3}$, Yanjie Fu$^{4}$, and Yan Solihin$^{\dagger 1}$}
\IEEEauthorblockA{$^{1}$University of Central Florida, FL, US \quad $^{2}$Universitas Gadjah Mada, Indonesia \\
$^{3}$Telkom University, Indonesia \quad $^{4}$Arizona State University, AZ, US \\
$^{\dagger}$Corresponding authors: \texttt{rian@ucf.edu, yan.solihin@ucf.edu}}
}


\maketitle

\IEEEpubidadjcol 

\def\thefootnote{$*$}\footnotetext{Much of the research was conducted when Santriaji was a postdoctoral researcher in the ARPERS group at UCF. \\ \\~\copyright 2025 IEEE.  Personal use of this material is permitted.  Permission from IEEE must be obtained for all other uses, in any current or future media, including reprinting/republishing this material for advertising or promotional purposes, creating new collective works, for resale or redistribution to servers or lists, or reuse of any copyrighted component of this work in other works.}
\def\thefootnote{\arabic{footnote}}

\begin{abstract}
Fault tolerance in Deep Neural Networks (DNNs) deployed on resource-constrained systems presents unique challenges for high-accuracy applications with strict timing requirements. Memory bit-flips can severely degrade DNN accuracy, while traditional protection approaches like Triple Modular Redundancy (TMR) often sacrifice accuracy to maintain reliability, creating a three-way dilemma between reliability, accuracy, and timeliness. We introduce NAPER, a novel protection approach that addresses this challenge through ensemble learning. Unlike conventional redundancy methods, NAPER employs heterogeneous model redundancy, where diverse models collectively achieve higher accuracy than any individual model. This is complemented by an efficient fault detection mechanism and a real-time scheduler that prioritizes meeting deadlines by intelligently scheduling recovery operations without interrupting inference. Our evaluations demonstrate NAPER's superiority: 40\% faster inference in both normal and fault conditions, maintained accuracy 4.2\% higher than TMR-based strategies, and guaranteed uninterrupted operation even during fault recovery. NAPER effectively balances the competing demands of accuracy, reliability, and timeliness in real-time DNN applications.\end{abstract}

\begin{IEEEkeywords}
Fault-tolerant Neural Networks, Memory Bit-flip, Resource-constrained system\end{IEEEkeywords}

\section{Introduction}
\label{sec:intro}

Fault tolerance in real-time systems with limited computational resources, or {\em resource-constrained} systems, presents significant integration challenges. These systems have finite computational capabilities that cannot be easily expanded to accommodate redundancy and recovery without substantial trade-offs. When deploying Deep Neural Network (DNN) in such systems, compromises in either timeliness or accuracy become inevitable due to computational limitations \cite{Zhu_2019_binaryensemble, mobilenetv3, tinybutaccurate, xu2023resilient}.

Consider an outdoor surveillance camera system operating in a tough environment, a typical soft real-time application that demands high availability despite occasional deadline misses being tolerable. These devices increasingly incorporate lightweight DNN models for real-time processing and privacy preservation \cite{wu2021pecam, guo2021mistify}. While they can tolerate some processing delays, their continuous operation is crucial. Importantly, due to operating in a tough environment (e.g., high temperature, high elevation), their memory chips are susceptible to bit flips—unintended changes in binary digit values within device memory—caused by environmental factors such as extreme temperatures, electromagnetic interference, and cosmic radiation \cite{hwang-asplos12-CosmicRay, electric_bitflip, baumann2005radiation}. Continuous operation of these devices in varying conditions may lead to hardware degradation and memory errors \cite{dutt2014aging}. This bit flip can have severe consequences. Research demonstrates that even a few bit flips in DNN weights can degrade model accuracy by up to 90\% \cite{siraj2022tbfa, DNN_bitflip_2, hanif2020dependabledl}, rendering these systems unreliable for their intended purposes.

There are two ways in which bit flips may affect computation accuracy or correctness. First, some systems may be deployed in environments exposed to higher bit flip rates than assumed by protection mechanisms, which might overwhelm them. Second, an even more serious problem is error accumulation. Even in environments with low bit flip rates, bit flips (if uncorrected) may accumulate, degrading accuracy over time as they accumulate~\cite{dutt2014aging, plettenberg2022cosmic}. Therefore, it is important for these systems to be augmented with {\em self-healing} ability, which includes error detection and error recovery without human intervention. Unfortunately, error detection or recovery from such bit flips can disrupt the timeliness of these DNN systems, resulting in delays in the inference. Addressing these memory bit flips in time-sensitive DNN resource-constrained systems to achieve desired accuracy and timeliness is challenging, bringing forth a three-way dilemma between DNN reliability, accuracy, and timeliness. 

Traditional approaches, like Triple Modular Redundancy (TMR), enhance reliability via model replication \cite{lyons1962use, liu2022using, ruospo2022selective,bertoa2022, cost-effective-iolts}. Despite their strengths, model replication reduces the resources available to each model, which in turn requires scaling down each model, compromising its accuracy. Conversely, DNNs with self-recovery \cite{ponader2021milr}, redirect all resources for recovery during fault. This "stop-the-world" recovery design makes the system unresponsive during recovery, likely causing missed deadlines, which is unsuitable for real-time applications.

To provide resiliency in a resource-constrained system without compromising accuracy much, we propose to use a {\em heterogeneous replication} approach via an {\em ensemble learning} \cite{sharkey1996combining}. Ensemble learning combines two or more model decisions to produce a higher accuracy than each base model alone. Ensemble learning works due to the significant diversity among the models, akin to how a multi-person jury tends to make a better decision than a single person. This paper leverages that principle by ensembling multiple resource-thin base models to provide resiliency while achieving a high accuracy comparable to a single large resource-rich model. In contrast, replicated models in TMR do not improve the combined model accuracy over individual models.

While an ensemble design was proposed to improve DNN robustness in prior works~\cite{ensemble,liu2019deep,wang2024garrison}, they ignore fault detection and self-healing. To address this limitation, we present NAPER (\underline{N}eur\underline{a}l Network \underline{P}rotection with \underline{E}nsemble \underline{R}edundancy). It introduces a key innovation through its novel integration of ensemble learning with error detection and a \textit{self-healing} mechanism. NAPER's error detection can detect any bit-flip affecting a single model. The \textit{self-healing} mechanism allows inferences to be made continuously while the affected model is recovering. They are then put together in a framework with a real-time scheduler that avoids missing deadlines by breaking down the recovery process into pieces (if necessary). Through this novel approach, NAPER simultaneously achieves high accuracy, self-healing, and timeliness.

Our evaluations under soft real-time constraints revealed that NAPER-protected models achieve about 40\% faster inference, both in bit-flip-free and when bit-flip is present, and 4.2\% higher accuracy compared to TMR.

\section{Related Works}
\label{sec:related}

DNN mitigation strategies in time-sensitive and resource-constrained environments must meet certain key criteria: \textbf{ (1) Accuracy}, self-healing support should minimally impact the system's accuracy. \textbf{(2) Availability}, The system should ensure that bit flip events do not hinder its capability to deliver a valid output within set deadlines, the maximum allowable time between input arrival and output generation for maintaining system functionality. \textbf{(3) Recovery}, The system should recover fully when faced with frequent or accumulating bit flips over time. \textbf{(4) Deployable}, The solution should be implementable without requiring custom hardware.

 Redundancy-based approaches (e.g., TMR) \cite{lyons1962use,ruospo2022selective,liu2022using,bertoa2022, cost-effective-iolts} offer model redundancy as their primary strength. While effective, it has significant computation and storage overheads that could compromise accuracy in resource-constrained environments. For instance, with a 200 ms deadline, unprotected ResNet can achieve nearly 75\% accuracy in CIFAR100. While the TMR-protected model can only achieve around 67\% accuracy. However, NAPER employs heterogeneous model redundancy to preserve DNN accuracy, achieving about 74\% accuracy.

Other works propose a self-recovery mechanism. MILR \cite{ponader2021milr} employs the algebraic properties of layer computation. While innovative, a key drawback of this strategy is its {\em stop-the-world} recovery approach, where recovery utilizes all computational resources, causing a temporary halt in the inference process. In contrast, NAPER's recovery can run {\em concurrently} with inference, ensuring minimal system downtime and providing uninterrupted service.

Another approach is to design a robust architecture resistant to a fault (Robust DNN). These can be broadly categorized into two types: ensemble-based methods \cite{ensemble,liu2019deep,wang2024garrison} that combine multiple models for enhanced reliability, and single-model approaches \cite{chen2022self,  wang2023aegis, kundu2024mendnet} that design a specialized architecture incorporate fault-resistant features. While these architectures provide robustness and maintain system availability, they lack the capability to recover from accumulated faults over time. The model will not be guaranteed 100\% the same as the model before bit flip, which may lead to significant accuracy degradation when bit flip accumulates or the fault rate is beyond its capability. Similarly, Error Correction Code (ECC)-based approach \cite{lee2022stealth, lee2022value, guan2019inplace} and some selective TMR approach \cite{ruospo2022selective, cost-effective-iolts}  have a similar issue on handling fault rate higher than its capability. In contrast, NAPER offers a comprehensive solution that maintains the model's accuracy and ensures its full recovery during faults.

While some approaches propose specialized hardware solutions like ReRAM (Resistive Random Access Memory) \cite{wang2022faulttolerant, shin2023faultfree, xia2017faulttolerant} to address fault tolerance, these solutions face significant deployment challenges. Such specialized hardware requirements, similar to ECC-based approaches \cite{lee2022stealth} that need ECC-capable DRAM or processors, limit their practical implementation. In contrast, NAPER's software-based solution offers a more accessible alternative, as it can be deployed on existing hardware infrastructure without specialized components.

Table~\ref{tab:bitflip_mitigation_methods} summarizes previous approaches and their limitations.  NAPER stands out by providing a solution that ensures accuracy, uninterrupted service, robust recovery, and adaptability to resource constraints.

\begin{table}[tbp]
\centering
\caption{Bit Flip Mitigation Approach Comparison}
\label{tab:bitflip_mitigation_methods}
\resizebox{\columnwidth}{!}{ 
\begin{tabular}{|l|c|c|c|c|}
\hline
\textbf{Approach} & \makecell[cc]{\textbf{Acc.}} & \makecell[cc]{\textbf{Avail.}} & \textbf{Recovery} & \textbf{Deployable} \\
\hline
\makecell[cl]{Redundancy-based \\ \cite{lyons1962use,ruospo2022selective,liu2022using,bertoa2022, cost-effective-iolts}} & - & \textbf{\checkmark} & \textbf{\checkmark} & \textbf{\checkmark} \\
\hline
Self-Recovery \cite{ponader2021milr} & \textbf{\checkmark} & - & \textbf{\checkmark} & \textbf{\checkmark} \\
\hline
\makecell[cl]{Robust DNN \\ \cite{chen2022self,liu2019deep,wang2024garrison,wang2023aegis,ensemble,kundu2024mendnet}} & - & \textbf{\checkmark} & - & \textbf{\checkmark} \\
\hline
\makecell[cl]{Hardware Support \\ \cite{wang2022faulttolerant,shin2023faultfree,xia2017faulttolerant}} & \textbf{\checkmark} & \textbf{\checkmark} & \textbf{\checkmark} & - \\
\hline
\makecell[cl]{ECC-based \\ \cite{lee2022stealth, lee2022value, guan2019inplace}} & \textbf{\checkmark} & \textbf{\checkmark} & - & - \\
\hline
NAPER & \textbf{\checkmark} & \textbf{\checkmark} & \textbf{\checkmark} & \textbf{\checkmark} \\
\hline
\end{tabular}
}
\end{table}

\section{Fault Model}
\label{sec:fault_models}

Our protection scheme addresses faults during inference runtime. In a neural network deployment, bit flip faults within memory-stored model parameters (e.g., weight and bias) critically impact the network's behavior and performance. These alterations can arise from various sources, including hardware faults, radiation, or electrical disturbances. Our scheme addresses transient and persistent bit flips affecting memory-stored model parameters within the memory hierarchy (caches and main memory). Protection against bit flip affecting logic circuits, code, or results of temporary calculation requires different schemes beyond this paper's scope. Similar to other redundancy-based schemes (e.g., TMR or DMR), NAPER can recover any number of bit flips as long as it only affects a single model on each layer.

\section{NAPER Design}
\label{sec:design}


NAPER encompasses three main components: (1) an ensemble redundancy design, (2) a dedicated fault detection and recovery module, and (3) a real-time scheduler.



\subsection{Model Ensemble Design}
\label{subsec:redundancy}

NAPER adopts heterogeneous redundancy models that have the same architecture but different parameter values. When combined as an ensemble, these models provide richer data representation and improved accuracy.  The ensemble's accuracy performance depends on model diversity: the more diverse these models are, the better the performance \cite{sharkey1996combining, lee2016stochastic, liu2019deep}. The simplest yet effective way to gain high accuracy is by employing independent training, training the new model independently from scratch with different random parameter initialization \cite{sharkey1996combining, lee2016stochastic}. \rian{Independent training can be challenging when training a single model is computationally intensive. Therefore, training redundancy independently will multiply the resource requirements}. Alternatively, fine-tuning or transfer learning \cite{yosinski2014transferable} or other ensemble generation schemes \cite{huang2016snapshot} can be used. However, our initial experiments showed that \textit{independent training achieves the highest accuracy}; therefore, NAPER primarily uses it for evaluation.

For the ensemble strategy, we use average voting instead of regular voting as NAPER assumes the user has a pre-trained base model. Average voting is more suitable for constrained environments due to lower computational overhead than other techniques \cite{wolpert1992stacked, sharkey1996combining}. Formally, given a base model $F_1(x;\Theta_1)$, where $\Theta$ is the model parameters and $x$ is the input; and $M$ redundant models $\{F_2(x;\Theta_2), \dots, F_{M}(x;\Theta_{M})\}$, the ensemble inference is $F_E(x; \Theta_E) = (\sum^{M}_{i=1} F_i(x;\Theta_i))/(M+1)$.


NAPER works with any number of redundant. In this paper, we utilize a single redundant model (\(M=1\)) to keep memory usage minimal, comparable with our baseline, TMR, while achieving adequate protection and accuracy improvement.

\subsection{Protection Design} 


In NAPER, we propose a novel relation parameters  $\delta_m^n$ on each layer to protect both the base model and redundant parameters. These parameters are computed by summing the parameters between the base model and its redundancy as shown in Eq. \ref{eq:delta}, where $\theta_m^n \in \Theta$ represents parameters in the \(n\)-th layer of $m$-th model, and  $\theta_1$ are parameters from the base model.   

\begin{equation}
    \label{eq:delta}
    \delta^n_m = \theta_1^n + \theta_m^n
\end{equation}

\subsubsection{\textbf{Fault Detection}}
NAPER employs two-step fault detection during inference for rapid processing, mainly when no errors occur in the model. The first fault detection step checks the validity of Eq. \ref{eq:delta} within a layer. This detection step provides fast error detection as we check two models simultaneously using only one comparison step compared to two comparisons in TMR. If no fault is detected,  this checking scheme could lead to faster inference. However, if there is a fault, this checking step does not identify the contributing model. Consequently, a secondary checking step is initiated for that specific layer. We use a checksum function by adding all parameters to check which model contributes to the fault.

\begin{figure}[tbp]
\centerline{\includegraphics[width=0.43\textwidth]{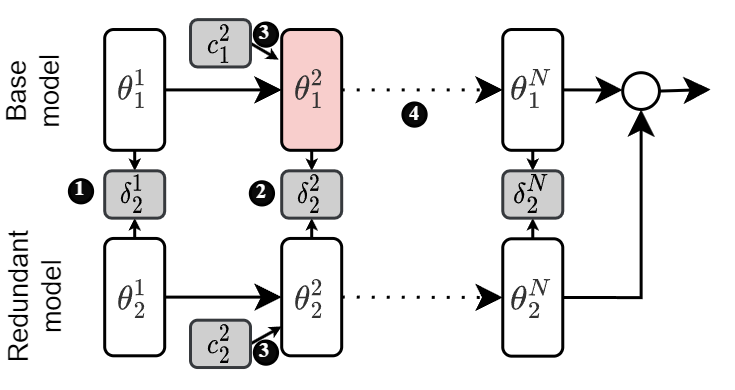}}
\caption{NAPER two-step detection scheme.}
\label{figdetect}
\end{figure}

We can see Figure \ref{figdetect} for illustration. There are two neural networks, with some bits flipped in the second layer of the first model ($\theta_1^2$). In the first layer, we only examined if Eq. \ref{eq:delta} still held \circled{1}. When a fault is observed in the first check of the second layer \circled{2}, we calculate the checksum for both the base and redundant models of the current layer's parameters and compare it to the pre-calculated checksum value, \(c^2_1\) and \(c^2_2\)  \circled{3}. When the checksum values are correct, we can assume there is a fault in \(\delta\). To ensure faster inference, when a fault is detected in one of the models, the inference computations on the following layers are stopped, but the detection keeps running to mark other faulted layers \circled{4}. 

\subsubsection{\textbf{Fault Recovery}}
Compared to prior robust ensemble DNN schemes \cite{ensemble,liu2019deep,wang2024garrison}, NAPER has a self-healing mechanism to recover the model from fault. NAPER protects base model parameters $\theta_1$, the redundant model parameters $\theta_m$, and the relation parameter $\delta_m$ as long as the fault only occurs in one of them in a layer. NAPER only needs to inverse Eq. \ref{eq:delta}, for \(\theta\) recovery, or recalculate it Eq. \ref{eq:delta} for \(\delta\) recovery. A layer recovery process is initiated with specific API calls instead of automatically triggered for flexibility in determining the timing for commencing the recovery process.

\begin{figure}[tbp]
\centerline{\includegraphics[width=0.37\textwidth]{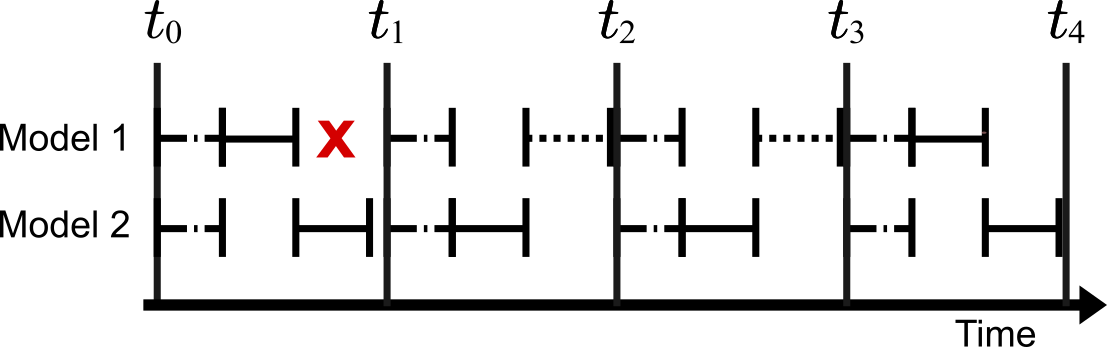}}
\caption{Visual explanation of NAPER's scheduling. Solid, dash-dotted, and dotted lines denote inference, fault detection, and recovery, respectively. A red X marks a bit flip event.}
\label{figrecoverytimeline}
\end{figure}

\subsection{Scheduler Design}

Let $I$ and $J$ represent the sets of models that are currently running and waiting to be run, respectively. Our scheduling strategy, designed for soft real-time systems, optimizes for two distinct operational scenarios: \textbf{(1)} On the system's operation without bit flips, we aim to optimize neural networks' expected inference accuracy \( \mathcal{A}(I, J) \). \textbf{(2)} When bit flips are encountered, the emphasis shifts to minimizing the time taken for system recovery, \( \Delta t_{rc} \), while ensuring the system maintains a minimum accuracy threshold, \( \mathcal{A}_{th} \). The scheduler tries to meet the above objectives with the following constraints: 

\begin{itemize}
    
    \item \textbf{Timing Constraint}: Given \( C_{in}\) and \( C_{dt} \), the observed worst-case execution time (WCET) for the inference and detection task, respectively, and \( \tau \), the time remaining until the deadline, we ensure \( \sum_{k \in J} (C_{dt_k} + C_{in_k}) \leq \tau \). This constraint ensures the system meets latency deadlines.
    
    \item \textbf{Accuracy Improvement Constraint}: Adding models from set \( J \) must result in a net positive gain in accuracy, \( \mathcal{A}(I) < \mathcal{A}(I, J) \), ensuring continuous  improvement. 
    
    \item \textbf{Threshold Constraint}: Given a predefined accuracy threshold, \(\mathcal{A}_{th} \), we ensure that the combined accuracy of the selected models \( \mathcal{A}(I, J) \geq \mathcal{A}_{th} \). To maintain a baseline quality of performance.
\end{itemize}

Figure \ref{figrecoverytimeline} visualizes NAPER's scheduler in action using one redundancy. Data batches arrive at timestamps $t_i$ and must be processed before $t_{i+1}$. During the detection stage, two parallel lines are shown because our detection scheme simultaneously checks two models. When a fault is detected before starting inference at $t_1$, the scheduler suspends operations for Model 1 and switches to alternative models to achieve the desired $\mathcal{A}_{th}$. With a short interval \( \tau \) before the arrival of the next data at $t_2$, NAPER commences Model 1 recovery operations and concurrently continues the recovery of the affected model if resources are available.

\subsection{Limitations and Discussion}

\textbf{Soft Real-time Guarantees.} Our model is premised on the characteristics of a soft real-time system, where occasional deadline misses are tolerable. However, frequent or prolonged delays pose risks, particularly in time-sensitive applications. Our observed worst-case execution time (WCET) is based on empirical data, not formal theoretical analysis.

\textbf{Handling Concurrent Bit Flips During Recovery.} If another bit flip occurs during recovery, NAPER remains resilient, continuing to recover the initial bit flip. Following this recovery, a fault check will always be conducted before inference of the incoming data, ensuring the integrity of DNN parameters before inference.

\textbf{Protecting Checksum Integrity.} To safeguard the integrity of the checksum, we can store it in a secure area of memory as in \cite{ponader2021milr}, or periodically refresh its value. Regular updates help ensure that the checksum remains reliable in detecting errors during operation.

\section{Evaluation Methodology}
\label{sec:methodology}

\subsection{Datasets and Baselines}
We evaluated NAPER's performance using CIFAR10, CIFAR100 \cite{Krizhevsky09learningmultiple}, and German Traffic Sign Recognition Benchmark (GTSRB) \cite{gtsrb} using ResNet models \cite{he2016deep, Idelbayev18a} and MobileNetV3 \cite{mobilenetv3}. We selected widely recognized models and datasets to easily compare with the prior works \cite{ ruospo2022selective}, and to aid in visualizing the significant impact of NAPER's accuracy enhancement.

We compare NAPER with five schemes:


\begin{itemize}
    \item \textbf{Triple modular redundancy (TMR)} \cite{lyons1962use}, a conservative approach that utilizes triple redundancies for each layer and a majority voting for detecting an error and recovering from it. We use TMR observed worst-case execution time (WCET) as the baseline in all experiments. 
    \item \textbf{CBR (Checksum-Based Recovery)}, detects faults by comparing layer checksums. We use the same checksum function with NAPER. It stores the model's healthy parameters on disk and retrieves them for recovery. 
    \item \textbf{DRO (Detection and Recovery Only)}, similar to TMR but only deals with memory bit flips like NAPER. Fault detection is achieved by comparing the parameters, and recovering by copying parameters from the other models.  
    \item  \textbf{Ensemble Fault Tolerance scheme (EFT)} \cite{ensemble}, is a fixed model ensemble of ResNet20, ResNet32, and ResNet44. EFT demonstrates an ensemble-based robust DNN design without re-training, fault detection, and recovery scheme.  
    \item \textbf{MILR} \cite{ponader2021milr}, a state-of-the-art self-healing that relies on algebraic relationships between layers of a neural network for layer recovery. 
\end{itemize}

\rian{For a fair comparison, all schemes above were implemented at the layer level: fault detection and recovery are conducted after each layer.} We exclude results of leveraging an ensemble in conservative methods, such as TMR, because these methods are not designed for ensemble strategy. Ensembling identical redundancy in TMR will not benefit the model's accuracy. We also exclude ECC-based protection schemes \cite{ lee2022value, lee2022stealth} as they are not designed to handle many bit flips. 

\subsection{Software and Hardware Specification}
NAPER\footnote{\url{https://github.com/ARPERS/NAPER}} is a wrapper for DNN models made with PyTorch. It was primarily tested on an NVIDIA Jetson Nano, a small computer module for edge AI applications. The Jetson Nano features a quad-core ARM Cortex-A57 CPU @ 1.43 GHz, a 128-core Maxwell GPU, and 4 GB of LPDDR4 RAM. 

\subsection{Bit flip Injection}

We use a bit flip fault injection scheme based on \cite{pytorchfi} and \cite{ares}, where we randomly flip bits in the memory blocks that store model parameters. The bit flips are uniformly distributed across the memory blocks with bit error rates ranging from $10^{-7}$ to $10^{-5}$, inline with prior works \cite{ ponader2021milr, mathew2018analysis}. In this paper, we refer \textbf{low fault rate} to the lowest bound ($10^{-7}$) and \textbf{high fault rate} to the highest bound ($10^{-5}$).

\subsection{Experiment Scenarios}

Our study focuses on three critical real-time DNN fault protection aspects: \textit{performance}, \textit{timeliness}, and \textit{accuracy}.

\textbf{Performance Evaluation:} We evaluated the additional overhead from NAPER and other methods during inference, either in bit-flip-free conditions or in the presence of bit flip. We simulate the bit flip events by injecting flips into model parameters on two settings: low fault rate and high fault rate. We conducted tests across various ResNet models on the CIFAR10 over 10 trials, averaged the results using the geometric mean, and normalized them to the TMR median.

\textbf{Timeliness Evaluation:} We conducted two evaluations. First, we simulated a video monitoring system and fed protected models with contiguous data at 4 frames per second (fps) for 50 frames. Then, we injected a high fault rate into model parameters five times randomly, \rian{demonstrating the accumulation of bit flips over time}. We chose 4 fps to meet TMR WCET. At the second evaluation, we varied the bit flip from low to high fault rate and measured the percentage of deadlines met. We define a model as meeting the deadline when it delivers the output before TMR WCET (ignoring accuracy). We used CIFAR10, averaged over 10 trials.

\textbf{Accuracy Evaluation:} We measure the model's accuracy when there are no bit flips and across different bit flip rates. An accuracy evaluation during bit flip was conducted using ResNet models on the CIFAR10 dataset at various fault rates ranging from low to high, averaged over 10 trials.

\section{Results and Discussions}
\label{sec:results}

\subsection{Performance Evaluation}

\begin{figure}[tbp]
\centerline{\includegraphics[width=0.5\textwidth]{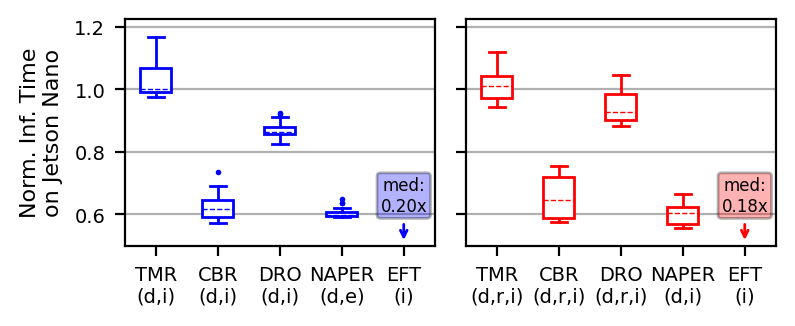}}
\caption{Normalized inference time to TMR of ResNet models in bit-flip-free (left) and bit flip conditions (right). "d", "i", "e", "r" denote bit flip detection, normal inference, ensemble inference, and recovery, respectively. Boxplots represent the 25th–75th percentiles. 'Med' denotes the median.}
\label{figitoff}
\end{figure}

In Fig. \ref{figitoff}-Left, we show inference times of various protection schemes, normalized to TMR's median, using different ResNet models for CIFAR10. Each scheme's task has bit flip detection (d) and inference (i), except  EFT scheme, which makes it achieve the fastest times. In our observation, fault detection can be up to 20x slower than fault recovery. TMR is the slowest because it needs two comparisons for fault detection (comparing between two redundancies), while NAPER detects faults 40.3\% faster with only one comparison. CBR is fast, relying only on checksum comparisons, but NAPER achieves higher accuracy with ensemble models. When faults occur, as shown in Fig. \ref{figitoff}-Right, TMR, CBR, and DRO execute recovery (r) during inference, adding overhead. CBR’s overhead varies due to parameter loading from the disk and the fault rate. NAPER prioritizes latency over recovery, using healthy redundant models for faster output, improving speed by 40.7\% over TMR.

\begin{table}[tbp]
    \centering
    \caption{Memory overhead (\%)}
    \footnotesize
    \begin{tabular}{|c|c|c|c|c|c|}
        \hline
        \multicolumn{2}{|c|}{\textbf{Scheme}}  & \makecell[cc]{\textbf{ResNet20} } & \makecell[cc]
        {\textbf{ResNet32} } & \makecell[cc]{\textbf{ResNet44} } & \makecell[cc]{\textbf{ResNet56} }\\
        \hline
        \multicolumn{2}{|c|}{TMR}  & 200.0 & 200.0 & 200.0 & 200.0  \\
        \hline
        \multicolumn{2}{|c|}{DRO} & 200.0 & 200.0 & 200.0 & 200.0 \\
        \hline
        \multicolumn{2}{|c|}{CBR} & 1.8 & 1.6 & 1.6 & 1.6 \\
        \hline
        \multicolumn{2}{|c|}{EFT} & 415.0 & 200.0 & 111.6 & N/A    \\
        \hline
        \multicolumn{2}{|c|}{NAPER ($M=1$)} & 201.0 & 200.8 & 200.8 & 200.8  \\
        \hline
    \end{tabular}
    \label{memoverhead}
\end{table}

\begin{figure}[tbp]
\centerline{\includegraphics[width=0.47\textwidth]{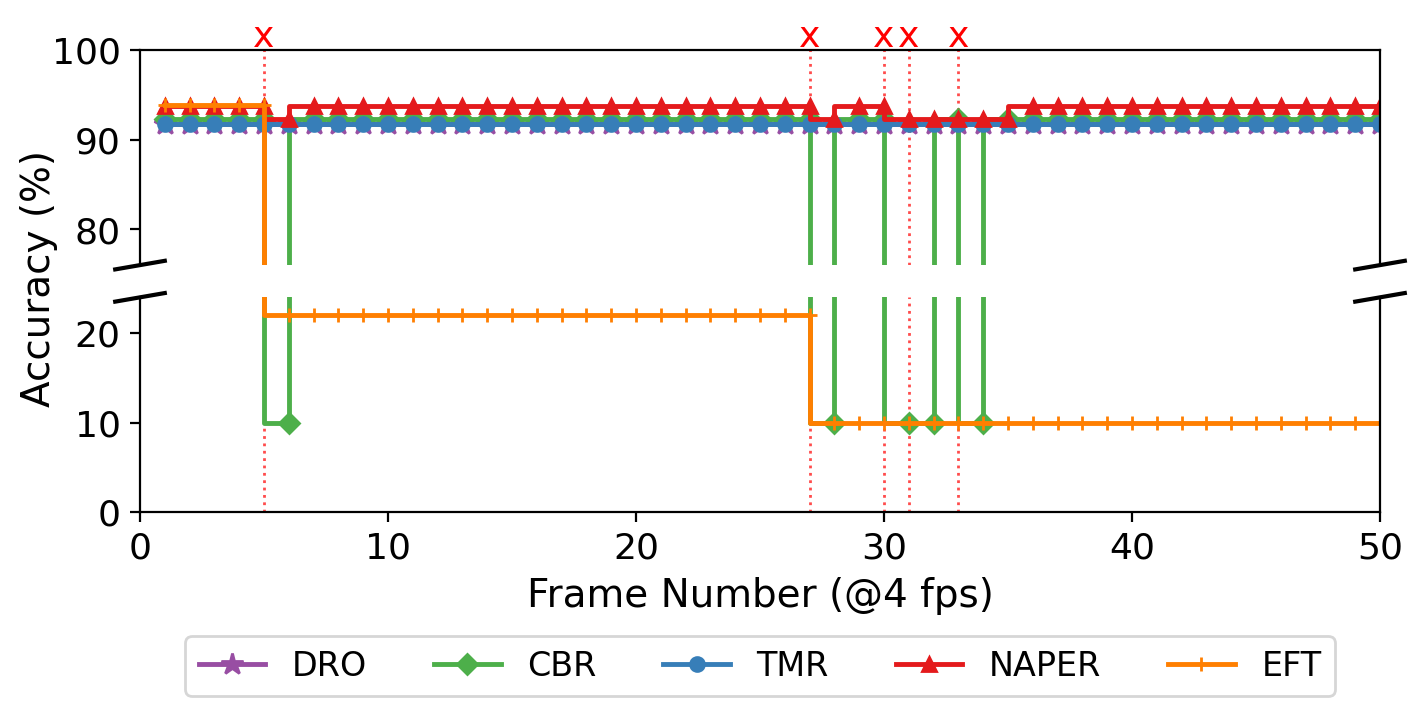}}
\caption{Evaluating protection schemes on consecutive data. A red X marks a moment when a bit flip occurs.}
\label{figsim}
\end{figure}

\begin{figure}[tbp]
\centerline{\includegraphics[width=0.37\textwidth]{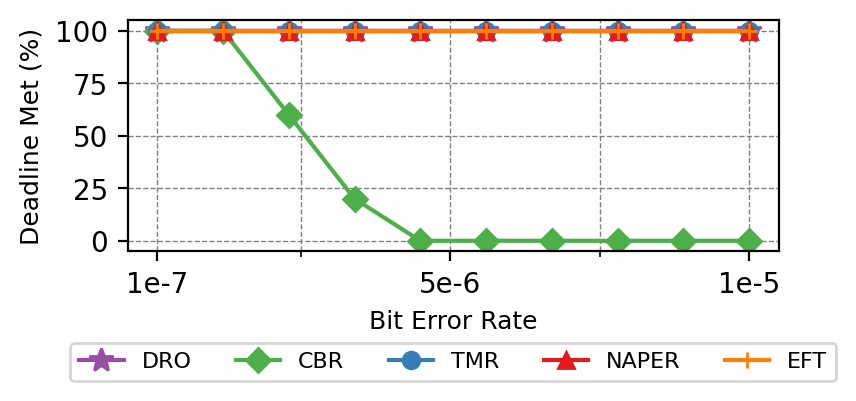}}
\caption{Average percentage of deadlines met using ResNet models on CIFAR10 when bit flips occur.}
\label{timelinessbitflip}
\end{figure}

The memory overheads for each scheme across different base models are shown in Table \ref{memoverhead}. TMR with two redundancies has 200\% memory overhead over a non-protected model. NAPER ($M=1$) adds negligible memory overheads compared to TMR for storing the checksum. CBR, with no redundancy, has the lowest memory usage. EFT, which has fixed models, can be seen from different perspectives, depending on the base model. If we define ResNet20 as the base model, then EFT has 415\% overhead, while if ResNet44 is the base, its overhead is smaller than TMR.

\subsection{Timeliness Evaluation}

We can see the results of the first assessment in Figure \ref{figsim}. TMR and DRO successfully met all the deadlines at four fps despite some bit flip events. However, TMR and DRO can only use the smallest model due to resource consumption, leading to the smallest model accuracy. NAPER, CBR, and EFT achieve better accuracy by using better models, as they need fewer resources. CBR has a slow recovery when bit flip occurs, making it miss some deadlines, indicated by an accuracy drop. Here, the bit flip does not make the accuracy drop to 10\%. This 10\% is a random guess in CIFAR10 as no output is produced. EFT got an accuracy drop caused by a bit flip over time. In contrast, NAPER still met deadlines using another healthy model and postponed the recovery. Its accuracy drops as NAPER can not leverage the ensemble at some points, but no deadline is missing. Figure \ref{timelinessbitflip} shows the result of the second evaluation. We can see that TMR, NAPER, DRO, and EFT consistently meet deadlines.  However, CBR occasionally misses them due to longer recovery times.

\subsection{Accuracy Evaluation}

\begin{figure}[tbp]
\centerline{\includegraphics[width=0.48\textwidth]{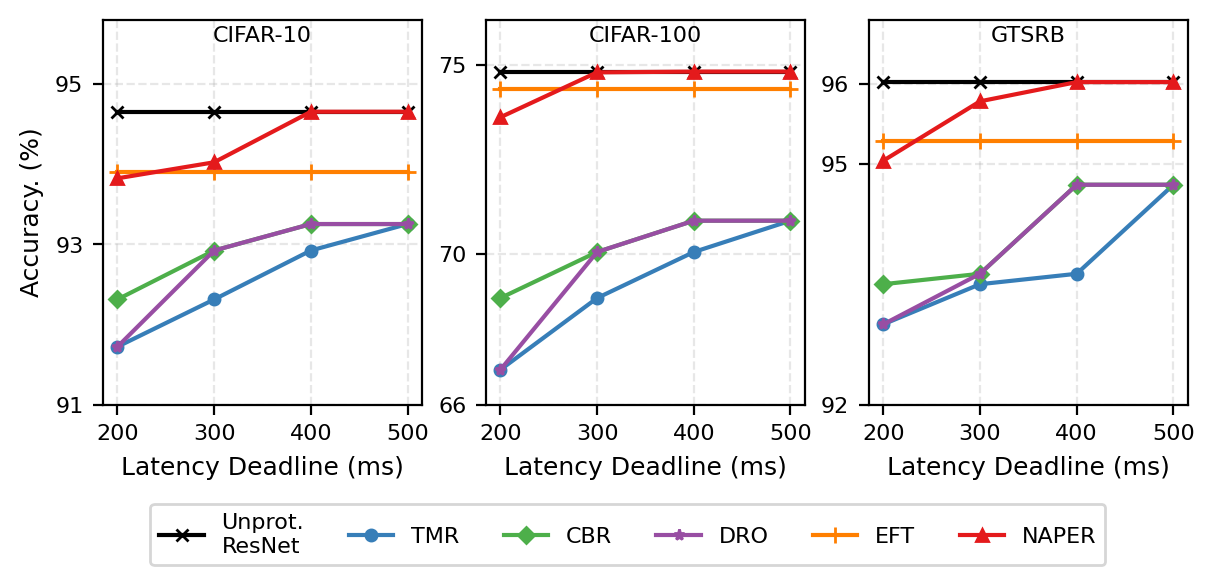}}
\caption{Accuracy achieved by various schemes across different latency deadlines using ResNet models.}
\label{accnobitflip}
\end{figure}

In bit-flip-free conditions, NAPER focuses on maximizing accuracy. Figure ~\ref{accnobitflip} shows the result accuracies of various schemes over deadlines. Each scheme utilizes the best model that still meets the deadline. For instance, in the lowest deadline in CIFAR10, TMR is only able to use the smallest model available and achieves the lowest accuracy at 91.7\%. With a 500-millisecond deadline, TMR utilizes the largest model available and achieves 93.3\%. However, it remains notably lower than the unprotected model. NAPER efficiently utilizes redundant models for ensembling, achieving higher accuracy across deadlines and datasets. Additional results using MobileNet can be seen in Table \ref{table:accnaper}. Here, our options are limited to only ensembling two models or not, as no larger models can fit the device. Averaging accuracy differences between NAPER and TMR across datasets and models in the lowest latency deadline, NAPER-protected models have $4.2\%$ higher accuracy compared to TMR.

\begin{table}[tbp]
    \centering
    \caption{Accuracy (\%) and average inference time (normalized to TMR) on MobileNet in bit-flip-free condition}
    \footnotesize
    \begin{tabular}{|c|c|c|c|c|}
        \hline
        \multicolumn{2}{|c|}{\textbf{Scheme}}  & \makecell[cc]{\textbf{CIFAR10} } & \makecell[cc]
        {\textbf{CIFAR100} } &\makecell[cc]{\textbf{Avg. Inf.}  \textbf{Time} } \\
        \hline
        \multicolumn{2}{|c|}{TMR}  & 86.2 & 57.1 & 1.0x  \\
        \hline
        \multicolumn{2}{|c|}{DRO} & 86.2 & 57.1 & 0.8x \\
        \hline
        \multicolumn{2}{|c|}{CBR} & 86.2 & 57.1 & \textbf{0.5x} \\
        \hline
        \multicolumn{2}{|c|}{NAPER} &\textbf{88.9} & \textbf{64.3} &\textbf{0.5x}  \\
        \hline
    \end{tabular}
    \label{table:accnaper}
\end{table}

\begin{figure}[tp]
\centerline{\includegraphics[width=0.37\textwidth]{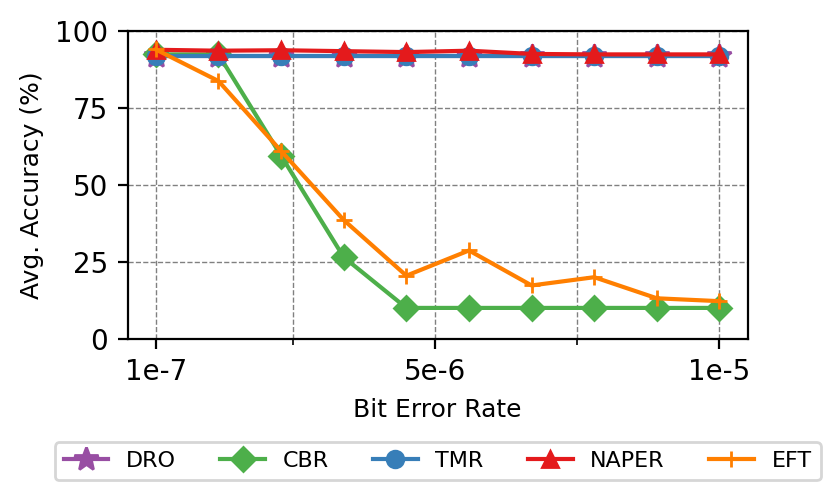}}
\caption{Average accuracy achieved on different deadlines using ResNet models on CIFAR10 when bitflip occurs.}
\label{accbitflip2}
\end{figure}

Figure \ref{accbitflip2} illustrates the impact of bit flip rates on accuracy. TMR and DRO maintain performance across fault rates, while CBR suffers from missed deadlines. EFT, while not missing deadlines, loses effectiveness with frequent faults. NAPER schedules recovery during slack time, ensuring deadline compliance, resulting in superior accuracy even under faults.

\subsection{Evaluating to Self-Recovery Scheme}
\label{sec:sensitivity_edge_cases}

\begin{figure}[tbp]
\centerline{\includegraphics[width=0.45\textwidth]{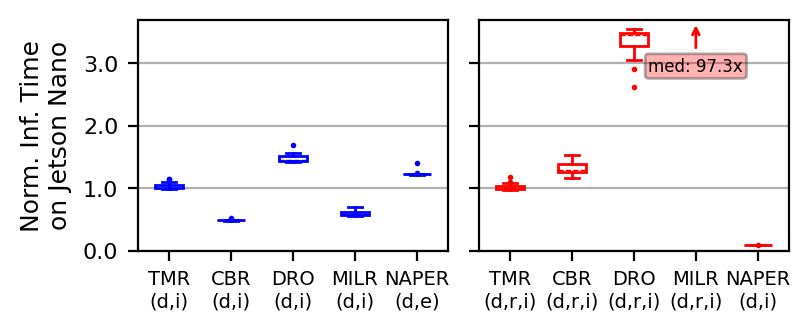}}
\caption{Inference time of different schemes normalized to TMR using a small neural network. Details consistent with Fig \ref{figitoff}.}
\label{figitoffmilr}
\end{figure}

We compare NAPER to MILR \cite{ponader2021milr} on CIFAR10 using a small two-layer neural network. We use a simple neural network model due to MILR's limitation: its protection scheme highly depends on layer computation. Implementing MILR in more complex models such as ResNet is not trivial. By using a significantly different model architecture, we can also assess NAPER's and other schemes' adaptability and performance on diverse neural network designs. In Figure \ref{figitoffmilr}-Left, NAPER's performance is slightly slower compared to the earlier results, due to the higher number of comparisons required for fault detection in this smaller model. However, NAPER's efficiency becomes more evident as the model complexity increases, such as when using ResNets and MobileNet. In Figure \ref{figitoffmilr}-Right, when bit-flip occurs, MILR's inference time is significantly impacted, potentially causing service interruptions. In contrast, NAPER maintains its performance and becomes the fastest approach, 90\% faster compared to TMR. This highlights NAPER's ability to handle faults efficiently, ensuring minimal impact on inference time and service continuity.

\section{Conclusion}
\label{sec:conclusion}

Integrating DNNs in real-time resource-constrained systems presents unique challenges, especially in the face of memory bit flips. We introduce NAPER, a novel solution that utilizes ensemble DNN models for enhanced accuracy and reliability. NAPER ensures accuracy, timeliness, and high system availability. Evaluations reveal NAPER's superiority in mitigating accuracy degradation compared to traditional strategies in resource-constrained settings and ensuring minimal system downtime during bit flip recovery.

\section{Acknowledgements}
This work was partially supported by the National Science Foundation under grant 2106629, Office of Naval Research under grant N00014-23-1-2136, and the Ministry of Education, Culture, Research, and Technology of the Republic of Indonesia through the Garuda Ace Project, with Project IDs 21-I1 and 24-I4.

\bibliographystyle{IEEEtran}
\bibliography{references}

\end{document}